\documentclass[runningheads]{llncs}
\usepackage{graphicx}
\usepackage{comment}
\usepackage{amsmath,amssymb} 
\usepackage{color}
\usepackage{ifthen}
\usepackage{epsfig}
\usepackage{array}
\usepackage{float}
\usepackage{subcaption}
\usepackage[font={small}]{caption}

\newcommand{\commentout}[1]{} 

\newcolumntype{C}[1]{>{\centering\let\newline\\\arraybackslash\hspace{0pt}}m{#1}}
\newcommand{\RNum}[1]{\uppercase\expandafter{\romannumeral #1\relax}}
\newcommand*{\affmark}[1][*]{\textsuperscript{#1}}

\begin{document}
\title{Human Motion Transfer from Poses in the Wild} 

\titlerunning{Human Motion Transfer from Poses in the Wild}
%
\author{Jian Ren\affmark[1] \and
Menglei Chai\affmark[1] \and
Sergey Tulyakov\affmark[1] \and Chen Fang\affmark[2] \and Xiaohui Shen\affmark[2] \and Jianchao Yang\affmark[2]}
%
\authorrunning{J. Ren et al.}
%

\institute{\affmark[1]Snap Inc., \affmark[2]ByteDance Inc.}

\maketitle              
%


\begin{figure}
\begin{center}
\centering
\includegraphics[width=1\linewidth]{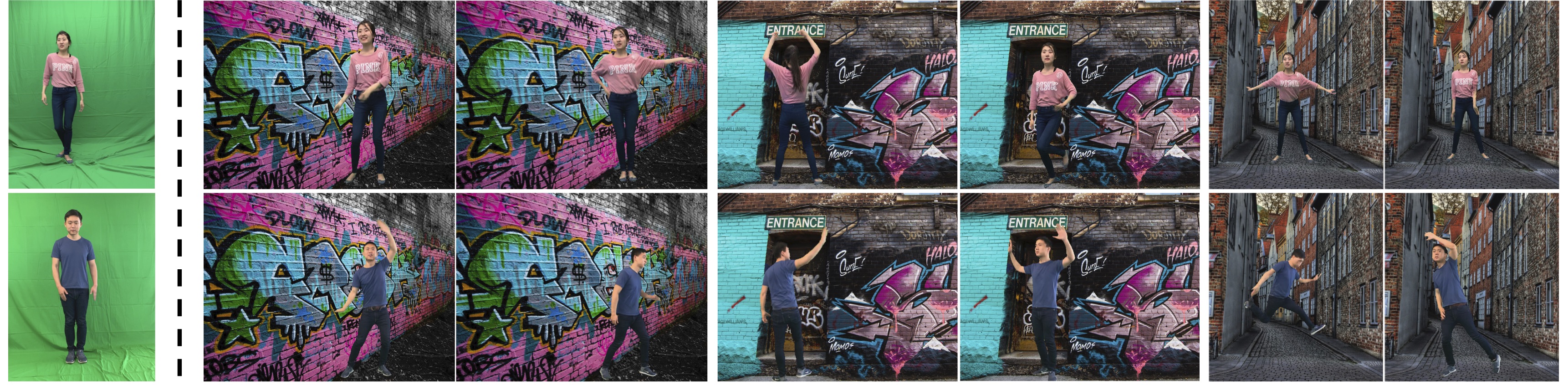}
\caption{\textbf{Application of our method}. We generate images of the target subject (left) and output foreground content only, allowing them to be easily blended with a new background (right).}\label{fig:cb}
\end{center}
\vspace{-3mm}
\end{figure}

\begin{abstract}
In this paper, we tackle the problem of human motion transfer, where we synthesize novel motion video for a target person that imitates the movement from a reference video. It is a video-to-video translation task in which the estimated poses are used to bridge two domains.
Despite substantial progress on the topic, there exist several problems with the previous methods. First, there is a domain gap between training and testing pose sequences--the model is tested on poses it has not seen during training, such as difficult dancing moves. Furthermore, pose detection errors are inevitable, making the job of the generator harder. Finally, generating realistic pixels from sparse poses is challenging in a single step.
To address these challenges, we introduce a novel pose-to-video translation framework for generating high-quality videos that are temporally coherent even for in-the-wild pose sequences unseen during training. We propose a pose augmentation method to minimize the training-test gap, a unified paired and unpaired learning strategy to improve the robustness to detection errors, and two-stage network architecture to achieve superior texture quality. To further boost research on the topic, we build two human motion datasets. Finally, we show the superiority of our approach over the state-of-the-art studies through extensive experiments and evaluations on different datasets.
\end{abstract}

\section{Introduction}
Video synthesis receives growing attention from research and industrial communities due to a wide range of applications. Among them, human motion retargeting saw significant progress, showing that by utilizing up-to-date deep neural network design and training techniques, approximate human motion can be transferred from one video to another. Such methods make it possible to generate a personalized dancing video of a subject not having any dancing experience. For example, one can animate themselves by using a ballerina video; or generate motion synchronized videos from multiple persons to be used for fake video detection~\cite{chan2019everybody}. 

\begin{figure}[t]
\begin{center}
\includegraphics[width=0.6\linewidth]{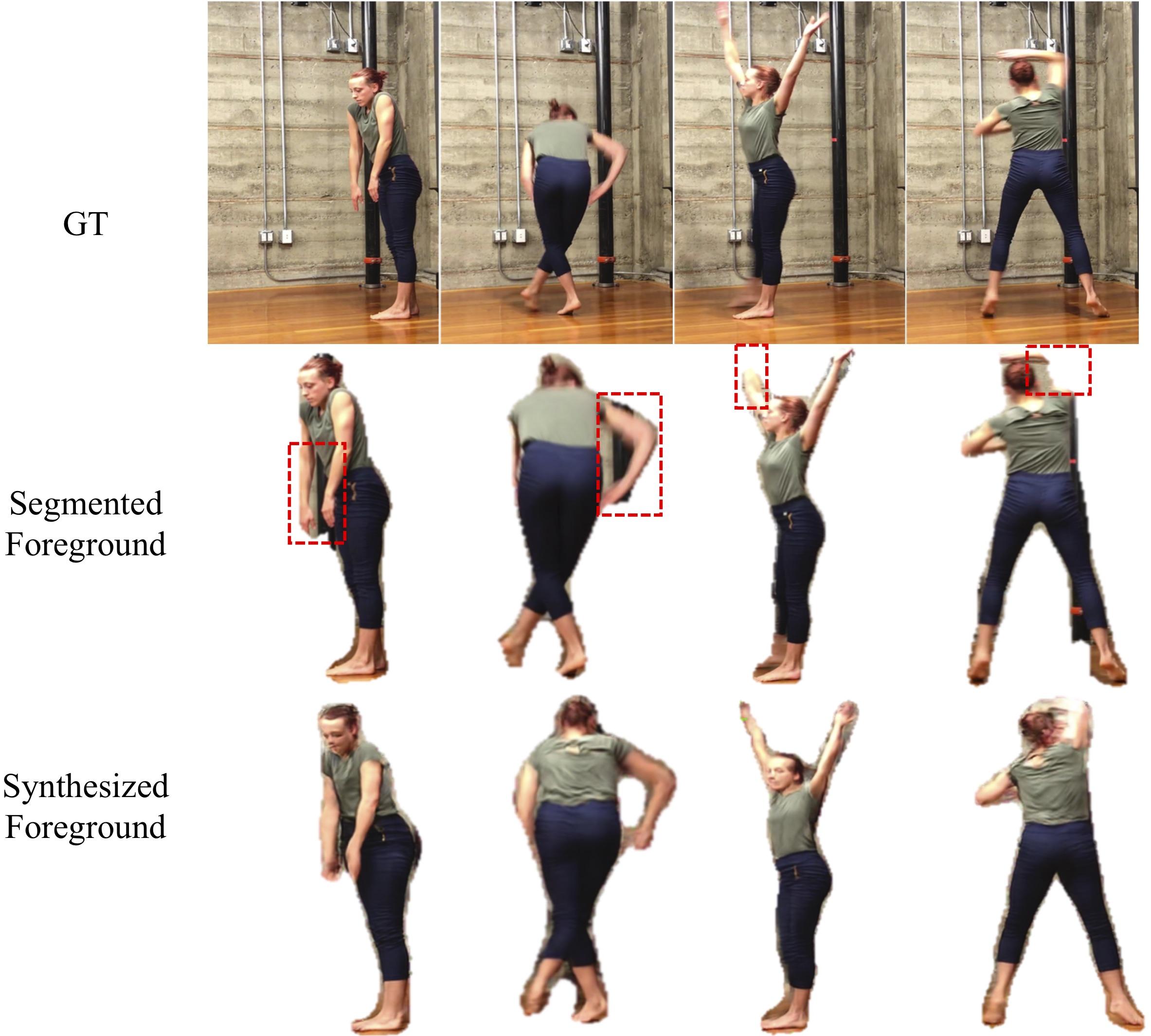}
\caption{Comparison of the segmented and synthesized foreground for the validation images. Top row: four ground-truth (GT) images; middle row: the segmented foreground from GT images using a state-of-the-art segmentation network~\cite{deeplabv3plus2018}; last row: the synthesized foreground using our methods. The segmented foreground images introduce extra background regions (left two columns) and miss parts of right arms (right two columns), as indicated by red boxes, while our methods can generate intact body images with less background region.}
\label{fig:edn}
\end{center}
\end{figure}

Similarly to the Everybody Dance Now work~\cite{chan2019everybody} and other video-to-video translation works~\cite{zhou2019dance,wang2018video}, our method requires a training video of a person performing a variety of motions. An off-the-shelf body pose detector~\cite{cao2018openpose} is used to parse pose skeleton and represent it as multi-channel pose maps to feed our network. Then, instead of focusing on generating the entire frame, as previous methods do~\cite{chan2019everybody,wang2018video}, we argue that using the foreground only (e.g., the person) improves performance and increases the number of possible applications (Figure~\ref{fig:cb}). Furthermore, focusing on foreground saves network capacity and computation time and costs, and allows the generated foreground to be easily reused on a new background.
Generating the entire frame limits the side movements of the generated person as the situation when the person leaves the known background region is not handled by traditional methods. We, therefore, focus on the foreground only. For the application of changing background, further limitations of generating the whole frame, such as introducing extra background region and missing body limbs, are presented in Figure~\ref{fig:edn}. 

Despite substantial progress, state-of-the-art human motion retargeting methods are still far from perfect, with several questions remain open. How to generalize to arbitrary in-the-wild reference motions, including extreme poses not seen during training? How to achieve robust results against possible pose detection errors? How to obtain realistic texture details while keeping temporal consistency and smoothness of the video? These challenging issues prevent us from crossing the realism gap and achieving higher visual fidelity. To answer these questions and tackle the human motion translation problem, we need a more robust method that generalizes well across different pose domains, produces high fidelity texture details, and features temporal consistency at the same time. In this work, we attempt to answer these questions by proposing a novel two-stage pose-to-video translation network employing a unified paired and unpaired learning framework.

Pose estimation networks often fail on in-the-wild input frames (Figure~\ref{fig:comp_vid2vid}), so that mismatching body parts appear. Moreover, for some poses, especially in the in-the-wild scenarios, several keypoints can be missed. To tackle these issues and in order to enrich the variance in the input, enhancing the robustness of the network to keypoint detector errors, we propose a body pose augmentation method. Specifically, we drop out random pose channels and resize lengths of certain body parts. Furthermore, we observe that direct synthesis of realistic textures from sparse pose representation is often challenging (see Figure~\ref{fig:ablation} for examples), we propose a refinement network to refine textures from images that are obtained from a pose-to-image translation network. Finally, there exists a gap between training and testing poses; in-the-wild poses during testing will be substantially different from the training poses sequences extracted from the person itself. This limitation is by design, and we would like to handle poses that the reference person is not able to do themselves, including difficult dancing moves. To bridge the gap, we introduce unpaired learning into the paired training pipeline improving the generalization of the system. To make unpaired learning feasible, we collect a large-scale single-person activities (SPA) dataset and use the dataset as the source of in-the-wild input to the training pipeline, leveraging it along with the supervised paired training branch by adopting the carefully designed combination of discriminators.


With our method, we can generate superior results featuring texture realism, motion smoothness and robustness to poses never seen during training. We extensively evaluate our system by conducting comparisons with state-of-the-art human motion retargeting techniques by reporting quantitative and qualitative metrics on three different datasets. The results support that our method significantly outperforms existing works by obtaining better numerical scores and achieving higher visual fidelity.

In summary, our major contributions are three-fold:
\begin{itemize}
    \item We propose a pose-to-video translation framework to generate high fidelity videos for unseen in-the-wild poses by minimizing the domain gaps between testing and training pose sequences. 
    \item We conduct extensive experiments and evaluations, both quantitatively and qualitatively, which demonstrate the significant advantage over state-of-the-art methods with superior result quality and generalization ability on unseen input poses. When presented with the results generated by our method vs. the ground truth examples, the users are often unable to tell which is real, preferring our method in 48.1\% of cases.
    \item We collect 
    two datasets: a high-quality indoor human video dataset containing all training sequences of the target persons used in this paper, recorded in front of a green backdrop screen, and a single-person activities dataset including large-scale in-the-wild pose sequences that are used during both training and evaluation.
\end{itemize}

\section{Related Work} 

\textbf{Image-to-Image Translation.}
With the recent development in conditional GANs~\cite{mirza2014conditional}, various inputs, including but not limited to classification categories~\cite{odena2017conditional} and images~\cite{liu2017unsupervised},  can be utilized to condition high fidelity image synthesis, besides noise in previous works~\cite{denton2015deep,radford2015unsupervised}. Image-to-image translation~\cite{isola2017image} uses paired training data to transfer an image from one domain to another and introduces an encoder-decoder architecture with skip connections. The architecture and training methods have been widely adopted in follow-up work. After that, pix2pixHD~\cite{wang2018pix2pixHD} further proposes a multi-scale generator and discriminator structure with residual-blocks to synthesize high-resolution photo-realistic images. However, obtaining paired data may still be prohibitively difficult for some tasks. Unsupervised image-to-image translation methods are presented to focus on minimizing domain gaps using unpaired image data~\cite{zhu2017unpaired,yi2017dualgan}. Nevertheless, compared with single images, video translation is usually more challenging since temporal consistency should also be considered as one vital factor besides image quality that impacts the final result quality.

\noindent\textbf{Video-to-Video Translation.}
Early efforts on non-conditional video synthesis typically convert a latent code into low-resolution short video frames via a recurrent neural network~\cite{vondrick2016generating,tulyakov2018mocogan,saito2017temporal}. These methods often suffer from low-quality  results, similar to the works on future frames prediction, that also use adversarial training and image reconstruction losses~\cite{walker2017pose,walker2016uncertain,finn2016unsupervised,denton2017unsupervised}. To generate photo-realistic videos and allow more fine-grain controls, the conditional video generation methods have shown great potential recently by using a sequence of conditioning inputs~\cite{gafni2019vid2game,wang2018video,huang2017real,chen2017coherent}, such as semantic segmentation maps~\cite{wang2018video}. A further group of work attempts to perform video-to-video translation requiring only a few images for the reference person. Despite substantial progress achieved recently~\cite{zakharov2019few,liu2019few,Siarohin_2019_NeurIPS,ha2019marionette,wiles2018x2face}, the generated human motion results are far from being realistic~\cite{lee2019metapix,wang2018fewshotvid2vid}.
Our work also falls into the scope of video-to-video translation, specifically focusing on the pose to video translation for unseen in-the-wild poses.

\noindent\textbf{Human Motion Transfer.}
Synthesizing novel views for human face and body has been studied extensively~\cite{liu2019liquid,bansal2018recycle,de2018semi,esser2018variational,joo2018generating,zanfir2018human,martin2018lookingood,esser2019unsupervised,lorenz2019unsupervised,kim2019unsupervised}. 
Some methods transfer facial expressions with parametric face models~\cite{elor2017bringingPortraits,thies2016face2face}. Extending such method to animating bodies is not trivial. Similar to the face re-animation method of Kim et al.~\cite{kim2018deep}, human body motion transfer can be achieved by adopting image generation neural networks.
Villegas et al.~\cite{villegas2018neural,villegas2017learning} use pose to predict future frames to synthesize a new human video.
Ma et al.~\cite{ma2018disentangled,ma2017pose} use a reference image to synthesize novel view given a target pose.
Siarohin et al.~\cite{siarohin2018deformable} improve the approach by proposing a deformable network architecture.
Balakrishnan et al.~\cite{balakrishnan2018synthesizing} segment human body parts according to the target pose and use a spatial transformation sub-module to synthesize unseen poses. Instead of predicting future frames or generating single frames from an input pose, we focus on high fidelity personalized video generation, where a network is specifically trained for each target person.

\noindent \textbf{Discussion.} Our focus is to synthesize personalized videos given arbitrary reference pose sequences, utilizing training videos for the target subject. We borrow the basic problem formulation from existing literature on pose to image generation~\cite{chan2019everybody,wang2018video,zhou2019dance,aberman2019deep}. However, compared with Everybody Dance Now (EDN) ~\cite{chan2019everybody}, Video-to-Video Synthesis (vid2vid)~\cite{wang2018video}, and Zhou et al.~\cite{zhou2019dance}, besides different methods on enforcing temporal coherence, we also adopt a two-stage translation and refinement network to improve generation texture quality, and propose a novel unified learning framework incorporating both paired and unpaired training examples to help the network generalize better on unseen in-the-wild poses.  We demonstrate the clear advantage of our methods over EDN~\cite{chan2019everybody} and vid2vid~\cite{wang2018video} in experiments.


\begin{figure}[t]
\begin{center}
\includegraphics[width=1\linewidth]{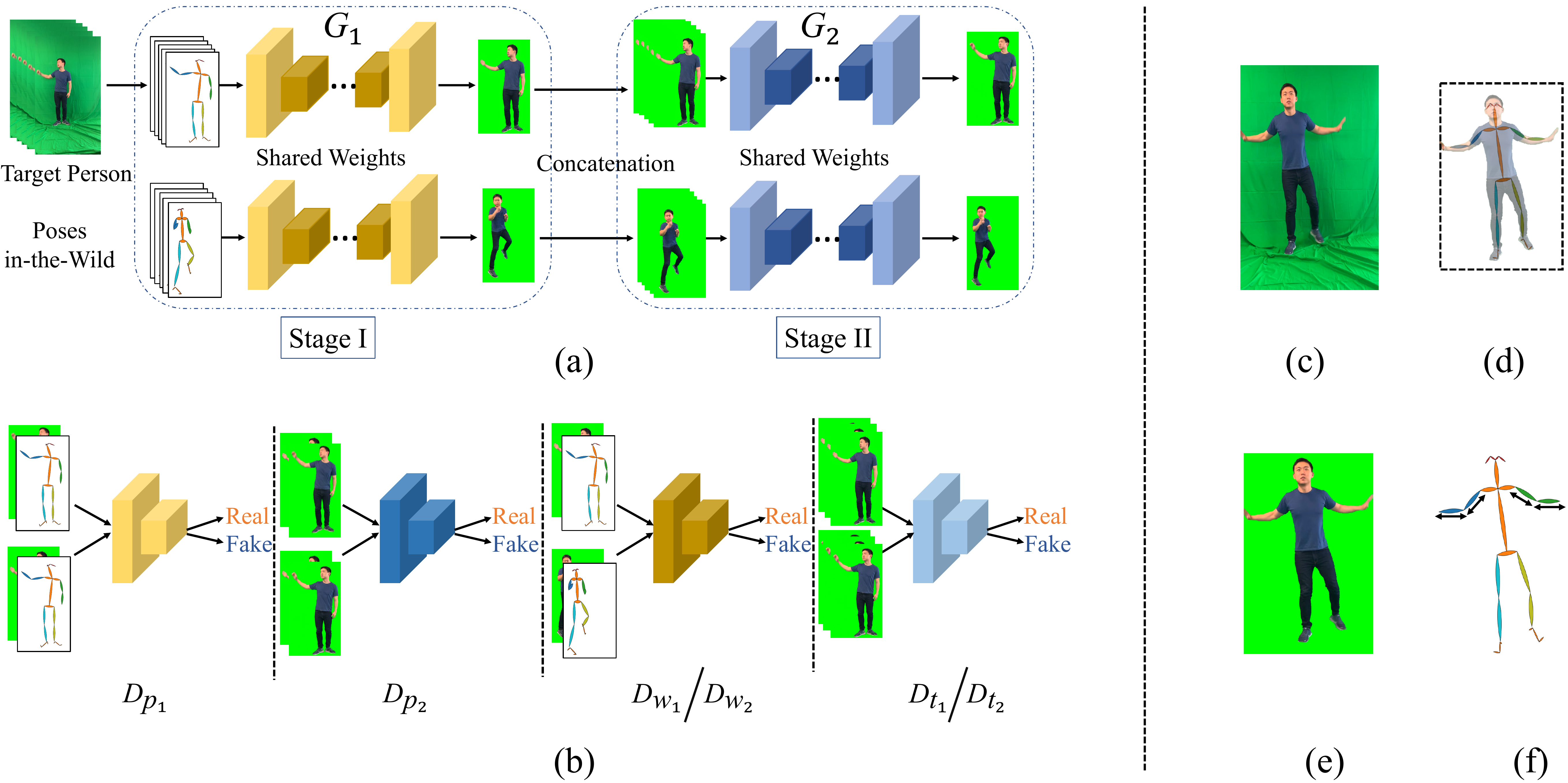}
\caption{\textbf{Pipeline and data augmentation}. (a) Our architecture consists of a Pose2Video network (Stage \RNum{1}) and a texture refinement network (Stage \RNum{2}). The top branch represents paired learning, and the lower one indicates unpaired learning. (b) Three types of discriminator: $D_p$ is the single frame discriminator for paired training, which is conditioned on the input to its corresponding stage (pose maps for $D_{p1}$ and Stage \RNum{1} output for $D_{p2}$); $D_w$ is the unpaired single frame discriminator, always conditioned on the input pose maps; $D_t$ is the temporal discriminator accepts stacked generated continuous frames. (c) One example frame captured in front of the green screen. 
(d) Extract the pose skeleton and remove the background, the pose information is further used to calculate the cropping region around the person. (e) Crop foreground and fill background with solid green color. (f) Perform data augmentation on the poses, such as elongating or shortening arms as indicated by arrows.
}
\label{fig:pipelinedata}
\end{center}
\end{figure}


\section{Method}
Given a source pose sequence $\{\mathbf{p}_1,\mathbf{p}_2,\ldots,\mathbf{p}_T\}$, our goal is to generate a photo-realistic human video sequence $\{\mathbf{f}_1,\mathbf{f}_2,\ldots,\mathbf{f}_T\}$ of the target person performing a sequence of motions provided by the input sequence of poses.

To train the method, for each subject, we record a set of video clips, covering common body poses and motions. We then parse the human pose skeleton from each video frame so that the paired training data is formed. Each pair includes a pose skeleton and the corresponding ground-truth image. In the setting of conditional GAN~\cite{mirza2014conditional}, a pose-to-image translation network can be used to synthesize a body video from input poses.

However, for such a challenging task, there are a few critical issues that prevent the baseline framework from working well in most general cases:
\begin{itemize}
\item There exist multiple domain gaps between the training and in-the-wild poses, including different recording devices/environments, subject identities, and motion styles. Direct inference on in-the-wild poses using networks trained purely with paired supervision will produce notably degraded results (Table~\ref{tab:AMT_wild});
\item State-of-the-art pose detection methods often lose track of important body key-points (top row in Figure~\ref{fig:comp_vid2vid}b) or mispredict (top row in Figure~\ref{fig:comp_vid2vid}c). Such noisy inputs make the generation even harder;
\item  A single-stage pose-to-image translation network tends to focus more on mapping the input pose to a rough spatial body image layout. Its limited capability is usually not sufficient to achieve high local visual realism at the same time, especially on dynamic texture details of body and clothes.
\end{itemize}
We detail our solutions to these problems in the next sections.


\subsection{Pose2Video Network}
As the first stage, the Pose2Video network performs pose-to-video translation, as shown in Figure~\ref{fig:pipelinedata}a ($G_1$ in Stage \RNum{1}). We encode the tracked skeleton with $n_s$ parts, such as arms and legs, into an input pose map with exactly $n_s$ channels by drawing a line segment of each part at its corresponding map channel.
The network structure follows pix2pixHD~\cite{wang2018pix2pixHD}. However, instead of doing image-to-image translation at the single-frame level, we adapt the network to accept multi-frame input to further utilize training sequence pairs for better temporal consistency and smoothness. For each training data pair ($\mathbf{p}_t, \mathbf{f}_t$) at time $t$, where $\mathbf{p}$ and $\mathbf{f}$ represent input pose maps and corresponding ground-truth frames respectively, we collect and stack $2K+1$ frames centered around $t$ as network input $\tilde{\mathbf{p}}\equiv\{\mathbf{p}_{t-K},\ldots,\mathbf{p}_{t+K}\}$, to help the network gain a better understanding of the temporal context. For discriminator $D_{p_1}$, we use the concatenation of pose and the ground-truth image as a true pair, and pose and the generated image as a false pair.

\noindent\textbf{Pose Augmentation}
The commonly used body tracking method~\cite{cao2018openpose,fang2017rmpe,xiu2018pose} may not perfectly accurate. Additionally, its performance drops substantially on the in-the-wild videos at inference time, which are usually more challenging comparing with our training sequences captured in a controlled environment. To remedy such an inevitable gap, we augment training pose maps by randomly dropping some input channels, perturb the location of joints keypoints, and elongate or shorten body part lengths in some channels, such as an example in Figure~\ref{fig:pipelinedata}f, so that both lengths and location of body limbs can be randomly changed during training. 

\subsection{Texture Refinement Network}
Our Pose2Video network can already generate video output conditioned by the input pose sequence. However, to tackle the challenging problem of photo-realistic human video generation, it is often difficult for the network to directly synthesize realistic texture from sparse pose in a single stage. Therefore, we concatenate an additional texture refinement network after Pose2Video, as shown in Figure~\ref{fig:pipelinedata} ($G_2$ in Stage \RNum{2}), to further refine the local texture details based on the rough output from the first stage. The texture refinement network also follows the setting of condition GAN  where the inputs for generator $G_2$ are adjacent $2K+1$ frames of Pose2Video RGB outputs $\tilde{G_1}(\tilde{\mathbf{p}})$, where  $\tilde{G_1}(\tilde{\mathbf{p}})\equiv\{G_1(\tilde{\mathbf{p}}_{t-K}),\ldots,G_1(\tilde{\mathbf{p}}_{t+K})\}$. For discriminator $D_{p_2}$, the concatenated  $\tilde{G_1}(\tilde{\mathbf{p}})$ and a ground-truth image is treated as true pair, and $\tilde{G_1}(\tilde{\mathbf{p}})$ and a generated image is false pair.


\subsection{Unified Paired and Unpaired Learning}
Even with our pose augmentation method, the paired poses extracted from the recorded training data may still have limited coverage over the vast human motion space, given that users may probably want to test challenging motion sequences at inference time such as ballet and hip-hop moves. Furthermore,  in-the-wild videos with complex background and occlusions also tend to raise the difficulty of pose estimation comparing to our chroma-keyed training sequences. In light of these, we propose to not only use augmented pairs of training data but also introduce unpaired learning with in-the-wild poses to boost the generalization ability and robustness.

We unify paired and unpaired training branches by swapping inputs and learning objectives while sharing weights of the target networks, as shown in Figure~\ref{fig:pipelinedata}a. In what follows, we introduce learning strategies of both paired and unpaired training branches respectively and conclude with the full training objective.

\paragraph{Notation}
Stage \RNum{1} and \RNum{2} share similar network structures and learning objectives. For the purpose of simplification, we use the same notation system to elaborate the training strategies for both stages within a single framework.
Besides the paired pose map $\mathbf{p}_t$ and ground-truth frame $\mathbf{f}_t$ that are consistent across the paper, we also uniformly denote the corresponding input and generator of each network stage as $\mathbf{x}_t$ and $G$, where $\mathbf{x}_t\equiv\tilde{\mathbf{p}}$ and $G\equiv G_1$ for the Pose2Video network (Stage \RNum{1}), and $\mathbf{x}_t\equiv \tilde{G_1}(\tilde{\mathbf{p}})$ and $G\equiv G_2$ for the refinement network (Stage \RNum{2}).

\paragraph{Paired Learning}
In the paired learning branch, the existence of ground-truth images allows us to enforce paired supervision on the network output. Instead of using low-level pixel objectives such as L1 reconstruction loss, we measure the perceptual similarity loss~\cite{johnson2016perceptual} $\mathcal{L}_{vgg}(G(\mathbf{x}_t), \mathbf{f}_t)$ with VGG19 network~\cite{simonyan2014very} between the network output and the corresponding ground-truth, to let the network gain a better semantic understanding and avoid blurriness caused by imperfect pose-image pairing and non-deterministic texture details regarding the pose.

We also use a single-frame discriminator $D_p$ conditioned on the input pose to enforce natural result and proper pose correspondences:
\begin{equation}
 \mathcal{L}_{D_p}=\mathbb{E}_{\mathbf{x},\mathbf{f}}[\log D_p(\mathbf{x}_t,\mathbf{f}_t)]+\mathbb{E}_{\mathbf{x}}[\log(1-D_p(\mathbf{x}_t,G(\tilde{\mathbf{x}}_t))],
\label{eq:g1}
\end{equation}
where $\tilde{\mathbf{x}}\equiv\{\mathbf{x}_{t-K},\ldots,\mathbf{x}_{t+K}\}$ is the temporal stacking of adjacent $2K+1$ frames centered around $t$, as described in the Pose2Video Network section.

To achieve more stable GAN training, we also adopted the discriminator feature matching loss~\cite{wang2018pix2pixHD} $\mathcal{L}_{f_p}(G,D_p)$.

Besides single-frame quality, the generated sequences should also be temporally consistent and retain plausible motion quality. Therefore, we use an additional unconditioned temporal discriminator $D_t$ to tell whether a continuous subset of $M$ result frames is realistic or not in the temporal domain:
\begin{equation}
 \mathcal{L}_{D_t}=\mathbb{E}_{\mathbf{f}}[\log D_{t}(\mathbf{f}_{t}^{t+M-1})]+\mathbb{E}_{\mathbf{x}}[\log(1-D_{t}(G(\tilde{\mathbf{x}})_t^{t+M-1})],
\label{eq:temporal}
\end{equation}
where $G(\tilde{\mathbf{x}})_t^{t+M-1}$ stands for stacking $M$ frames of generator outputs \\ $\{G(\tilde{\mathbf{x}}_t),\ldots,G(\tilde{\mathbf{x}}_{t+M-1})\}$. We also use a feature matching loss  $\mathcal{L}_{f_t}(G,D_t)$ for $D_t$.

\paragraph{Unpaired Learning}
During unpaired learning, instead of using recorded paired data, we randomly feed the network with body pose inputs extracted from video sequences with different subjects. Incorporating these in-the-wild inputs helps bridge the pose domain gap and increase network robustness against unseen inputs.
Without paired ground-truth supervision, we perform unpaired training based on network models that share weights with paired training and solely adopt a single-frame discriminator $D_w$ similar to that introduced in the paired learning section. Different from $D_p$, those positive examples do not share the same condition $\mathbf{p}_t$ as the negative ones, but randomly draw pairs ${(\mathbf{p}'_t, \mathbf{f}'_t})$ from recorded sequences used in paired learning:
\begin{equation}
 \mathcal{L}_{D_w}=\mathbb{E}_{\mathbf{p},\mathbf{f}}[\log D_w(\mathbf{p}'_t,\mathbf{f}'_t)]+\mathbb{E}_{\mathbf{x}, \mathbf{p}}[\log(1-D_w(\mathbf{p}_t,G(\tilde{\mathbf{x}}_t))].
\label{eq:g1}
\end{equation}

\paragraph{Full Objective}
We train our networks with a two-stage training strategy that we first train the Pose2Video network $G_1$ by optimizing $D_{p1}$, $D_{t_1}$, and $D_{w_1}$, and then fix the weights in $G_1$ and train the refinement network $G_2$ by optimizing $D_{p_2}$, $D_{t_2}$, and  $D_{w_2}$. The overall loss functions for both training stages are similarly defined as:
\begin{equation}
\small
\begin{split}
&\min\limits_G(\max\limits_{D_p}\mathcal{L}_{D_p}(G,D_p)+\max\limits_{D_t}\mathcal{L}_{D_t}(G,D_t)+\max\limits_{D_w}\mathcal{L}_{D_w}(G,D_w))+ \\
&\lambda_{vgg}\mathcal{L}_{vgg}(G)+\lambda_{fm}(\mathcal{L}_{f_p}(G,D_p)+\mathcal{L}_{f_t}(G,D_t)).
\end{split}\label{eq:stage1}
\end{equation}

\section{Experiments}

In this section, we conduct both quantitative and qualitative experiments to demonstrate the advantages of our method, especially on challenging poses.

\subsection{Experiment Setup}

\subsubsection{Data Preparation.}
We use three datasets to perform validation. The first one is released with the Everybody Dance Now (EDN) paper ~\cite{chan2019everybody}. The dataset includes videos of five subjects. We follow the training and validation strategy in~\cite{chan2019everybody} to perform experiments.  To facilitate the following presentation, we denote the validation videos from EDN as \textbf{EDN-Vali}.

For the second dataset, we collected videos containing four subjects. All videos are filmed in an indoor environment with the subject standing in front of a backdrop green-screen (as an example frame shown in Figure~\ref{fig:pipelinedata}(c)) to help isolate the foreground and achieve better segmentation quality. An iPhone fixed on a tripod is used to shoot the videos. During the process, all subjects are asked to either perform slow random moves or follow simple online dancing videos (these guidance videos are not used in either training or validation). 
On average, we collect $22$ minutes video for each target person.
We split all videos of each subject into training and validation sets with a ratio of $17:3$. We denote the validation set as \textbf{Target-Vali}. 

We process these captured data by first applying an off-the-shelf pose detection networks~\cite{cao2018openpose} to get estimated body poses. 
Then we perform chroma-key composition to mask out the target person and change the background to a solid green color. Finally, we crop each frame with the smallest rectangle that encloses the target person, as shown in Figure~\ref{fig:pipelinedata}(e).

Besides the two datasets for paired learning, we create a large-scale single-person activities dataset (\textbf{SPA}) for unpaired learning and validation.
To make SPA suitable for poses-to-video generation, we collect $1,060$ single-person activity videos and make sure that all these videos catch the whole human body, and each video only includes one person. The average duration of each video is about $10$ seconds and we extract frames at $30$ FPS, giving $315,000$ frames in total. The body poses of each SPA video are detected for training and inference uses.
Additionally, we randomly take out $64$ videos from SPA as a validation dataset, denoted as \textbf{SPA-Vali}, to verify model performance on unseen in-the-wild poses. The average duration of SPA-Vali videos is about $17$ seconds. Compared with the other two validation datasets, SPA-Vali is more challenging. 

\subsubsection{Implementation Details.}
We adopt the multi-scale generator and discriminator architecture and apply the progressive training schedule. We first train a model for 128$\times$256 and then upsample to 256$\times$512. We set $K=2$ for input pose maps $\tilde{\mathbf{p}}$, and $M=3$ in Eqn.~\ref{eq:temporal}. The hyper-parameters in Eqn.~\ref{eq:stage1}  are set as $\lambda_{fm}=10$ and $\lambda_{vgg}=10$.
We use the initial learning rate as 0.0002 and gradually decrease it.

\subsubsection{Evaluation Metrics.}
We numerically evaluate the generated videos with both objective analysis and subjective user studies.
\begin{itemize}
\item \textbf{Objective Metrics.}
We adopt the three widely-used objective metrics to assess the result quality: SSIM (Structural Similarity)~\cite{wang2004image} index to measure the perceived image quality degradation between both synthesized and real frames; LPIPS (Learned Perceptual Image Patch Similarity)~\cite{zhang2018unreasonable} to measure the perceptual similarity between generated and real images; and FID (Fréchet Inception Distance)~\cite{heusel2017gans} to measure the distribution distance.
\item \textbf{Subjective Scores.}
We also conduct user studies to analyze the video quality regarding real human perception. The experiments are performed using the Amazon Mechanical Turk (AMT) platform. We design two settings for users to compare video quality: 1) Pairwise comparison: we show workers pairs of videos with exactly the same motion but from two different sources (including ground-truth, results of our method, or results of state-of-the-art methods), and ask them to choose the more realistic one. The two videos are shown side-by-side and their orders are randomly chosen; 2) Single-video evaluation: we only show workers a single video and ask them whether this video looks real to them. Workers can only choose a video as real or not.
\end{itemize}

\subsection{Comparison Results}
Both EDN~\cite{chan2019everybody} and vid2vid~\cite{wang2018video}  are state-of-the-art methods on human body motion transfer and have achieved significantly better results compared with existing image-based human body generation methods~\cite{balakrishnan2018synthesizing,wang2018pix2pixHD}. In the following sections, we compare our method with the two studies and show both qualitative and quantitative results.


\begin{table}[t]
\small
\begin{minipage}{.48\linewidth}
\centering
\caption{\textbf{Objective comparisons} on the EDN-Vali dataset.}
\begin{tabular}{C{2.8cm} | C{1.2cm}|  C{1.3cm} }
\hline
&SSIM $\uparrow$ &  LPIPS $\downarrow$  \\ \hline\hline
EDN & 0.838  &0.050\\ \hline
ours w/ BG & 0.948  & 0.027 \\ 
ours segmented FG &0.959 & 0.031\\
ours w/o BG & \textbf{0.976}  & \textbf{0.015} \\  \hline
\end{tabular}
\label{tab:EDN}
\end{minipage}%
\hfill
\begin{minipage}{.45\linewidth}
\centering
\caption{\textbf{Objective comparisons} on the Target-Vali dataset.}
\begin{tabular}{C{2cm} | C{1.2cm}|  C{1.2cm} }
\hline
&SSIM $\uparrow$ &  FID $\downarrow$  \\ \hline\hline
vid2vid & 0.9518  &7.5214\\ 
ours& 0.9666  & 5.9453 \\ \hline
\end{tabular}
\label{tab:SOTA}
\end{minipage} 
\end{table}

\subsubsection{Comparison with EDN~\cite{chan2019everybody}.} 
We use the data collected by EDN to train our models, and then validate the models on EDN-Vali dataset. We conduct three experimental settings: 1) \textbf{ours w/ BG:} similarly to EDN, we use original images with background for training and validation; 2) \textbf{ours segmented FG:} similarly to EDN, we use original images to train the networks, but apply a segmentation network~\cite{deeplabv3plus2018} on the generated images to get foreground region; 3) \textbf{ours w/o BG:} we apply a segmentation network on the training data and only use the segmented foreground to train our model, so the synthesized results only contain foreground. 
Following EDN, we run the experiments for five subjects and report the averaged results. 
The results in Table~\ref{tab:EDN} show that our method outperforms EDN significantly as we achieve much higher SSIM and much lower LPIPS. We also notice that using the foreground region for training performs better than using a whole frame as the networks can focus on synthesizing only the foreground part. Although the foreground region can be obtained by segmenting the generated image, the image quality is inferior to the foreground that is synthesized directly, supporting that it is beneficial to remove the background prior to training.

\begin{table}[t]
\small
\begin{minipage}{.55\linewidth}
\centering
\caption{\textbf{User study on paired comparison}. Each user is presented with a pair of videos generated by different methods and asked to pick the relatively better one.}
\begin{tabular}{C{1.85cm} | C{2cm} | C{2cm}}
\hline
& Target-Vali & SPA-Vali  \\ \hline\hline
vid2vid/ours  &  29.2\%/\textbf{70.8\%} & 9.7\%/\textbf{90.3\%} \\ \hline
vid2vid/GT& 36.2\%/\textbf{63.8\%} & - \\ 
ours/GT & 48.1\%/\textbf{51.9\%} &- \\ 
\hline
\end{tabular}
\label{tab:AMT_target}
\end{minipage}%
\hfill
\begin{minipage}{.4\linewidth}
\centering
\caption{\textbf{User study on single video evaluation}. Each user is presented with a single video and asked to choose if it is real.}
\begin{tabular}{C{1.1cm} | C{1.6cm}| C{1.6cm}}
\hline
&Target-Vali  & SPA-Vali  \\ \hline\hline
vid2vid &42.3\% &25.7\% \\ 
ours & 71.4\%& \textbf{44.5\%} \\ 
GT & \textbf{85.1\%} & -\\ \hline 
\end{tabular}
\label{tab:AMT_wild}
\end{minipage} 
\end{table}

\subsubsection{Comparison with vid2vid~\cite{wang2018video}.}
In their original article, vid2vid extracts human poses with both OpenPose~\cite{cao2018openpose} and DensePose~\cite{alp2018densepose}. However, DensePose detects human body shapes together with skeleton poses. The shape information encodes the identity of the reference subject, which can cause difficulty preserving the target identity during inference. 
In order to have a fair comparison, we implement their method using the same pose generation and augmentation methods as ours.


\begin{figure}[t]
\begin{center}
\includegraphics[width=1\linewidth]{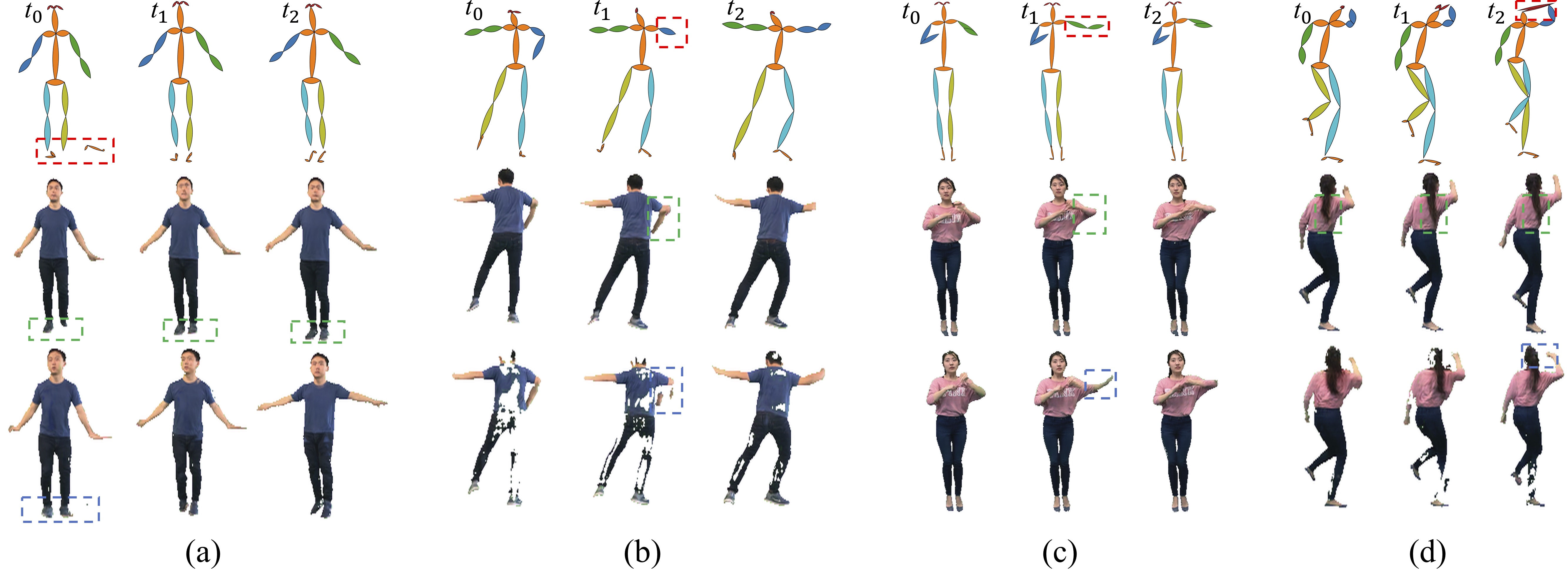}
\caption{\textbf{Visual comparisons with vid2vid~\cite{wang2018video}}. Each of these four sequences contains three consecutive frames, with the first row shows the input skeletons, the seconds row shows our results, and the third row shows the results by vid2vid. We emphasis the incorrect pose regions with red dashed box, and the corresponding results generated by our method and vid2vid with green and blue boxes respectively.}
\label{fig:comp_vid2vid}
\end{center}
\end{figure}

\paragraph{Quantitative Evaluations.} We list the average SSIM and FID metrics in Table~\ref{tab:SOTA}, evaluated on our Target-Vali dataset in which we have access to ground-truth. Compared with vid2vid~\cite{wang2018video}, we achieve noticeably higher scores on SSIM and lower scores on FID, which proves that our method can synthesize results with better objective quality.

We also present user study results, with both pair-wise comparison and single-video evaluation settings shown in Table~\ref{tab:AMT_target} and Table~\ref{tab:AMT_wild}. From the pair-wise comparison result, we can see that $70.8\%$ and $90.3\%$ users prefer our results to those by vid2vid on Target-Vali and SPA-Vali datasets respectively. Also, users demonstrate very close preferences between our results and the ground-truth ($48.1\%$ vs. $51.9\%$), which shows that our results are somehow comparable to the real videos. 
As for the single-video evaluation, we show the number of percentage videos that are rated as real videos by users. We can notice our method achieves better scores than vid2vid on both datasets as well. As expected, both methods receive lower scores on SPA-Vali compared with Target-Valid since SPA contains more challenging poses. However, our method continues to perform substantially better than vid2vid, demonstrating superior generalization.

\paragraph{Qualitative Evaluations.} For visual comparison, we randomly selected pose sequences from SPA-Vali and generate motion transfer results with both our method and vid2vid in Figure~\ref{fig:comp_vid2vid}. The results confirm that our approach generated more realistic results in cases when the pose detector fails to reliably find the keypoints.
For example, in top row of Figure~\ref{fig:comp_vid2vid}a, the left foot is incorrectly detached from the leg in the first frame. However, our method is still capable of generating the foot in the right position with consistent orientation, while vid2vid generates an unnatural dot at the wrong location, misled by the input pose.
In Figure~\ref{fig:comp_vid2vid}b, the right lower arm is entirely missing in the second frame. Our method successfully predicts the missing arm utilizing the adjacent frames, achieving significantly better results than vid2vid.
In Figure~\ref{fig:comp_vid2vid}c, the position of the left lower arm in the second frame is inconsistent with its neighbors. Our method again generates consistent results while vid2vid generates an extra arm at that frame.
In Figure~\ref{fig:comp_vid2vid}d, the detected face key-points are unnaturally stretched in the third frame. In contrast to the broken head result by vid2vid, our method still manages to generate intact head and hair with consistent direction.
Besides robustness, it is worth noticing that vid2vid often produce less satisfactory foreground masks, which leads to internal holes or missing parts as shown in Figure~\ref{fig:comp_vid2vid}b and Figure~\ref{fig:comp_vid2vid}d. As can be seen, our method consistently produces complete and accurate foreground boundaries.

\begin{table}[t]
\centering
\caption{\textbf{Ablation analysis on our method}. Our full method (PL-UL-Stage2) achieves significantly better results comparing with other alternatives.}
\begin{tabular}{C{2.5cm} | C{2.2cm}|  C{2.2cm}}
\hline
&SSIM $\uparrow$ &  FID $\downarrow$    \\ \hline\hline
PL-Stage1  & 0.9611 &8.0105\\
PL-Stage1-DA   &0.9633  &7.1197\\
PL-Stage1-DA-F   & 0.9651 &6.5701 \\
PL-Stage2  &0.9655  &6.2046 \\
PL-UL-Stage2 & \textbf{0.9666}  &\textbf{5.9453}  \\ \hline
\end{tabular}
\label{tab:ablation}
\end{table}

\begin{figure}[t]
\begin{center}
\includegraphics[width=0.8\linewidth]{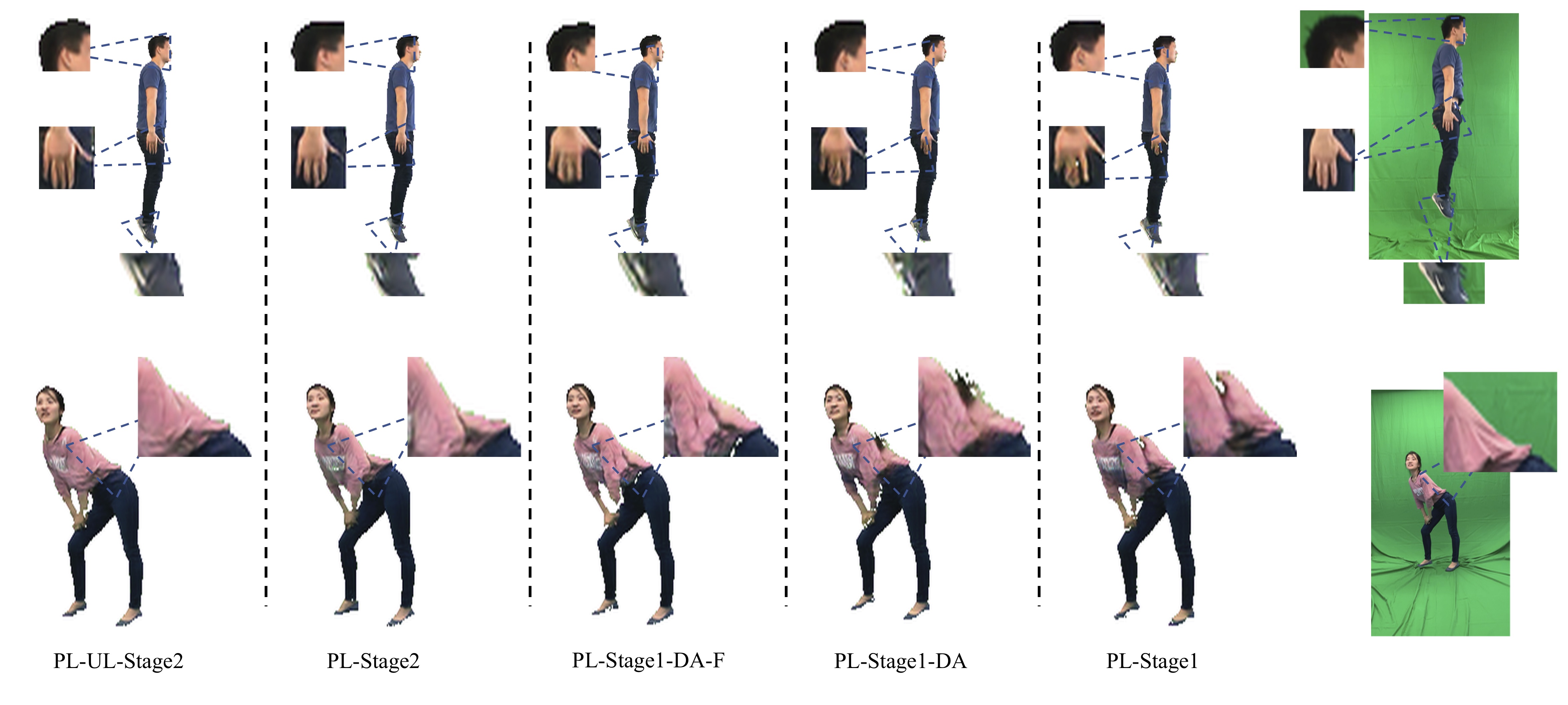}
\caption{\textbf{Qualitative ablation study results}. The result generated by our full method is shown on the leftmost, four results generated by alternative strategies shown in the middle with key components gradually removed at each step, and the ground-truth shown on the rightmost. We also show some zoom-in insets to better visualize the differences.}
\label{fig:ablation}
\end{center}
\end{figure}

\subsection{Ablation Analysis}
We perform ablation studies to identify which of the contributions are responsible for superior quality. We report the following experiments: 1) \textbf{PL-Stage1}: as the baseline, we adopt a single-stage Pose2Video network without unpaired learning and pose augmentation, and stack the input poses as $\{\mathbf{p}_{t-K},\ldots,\mathbf{p}_{t}\}$, which follows the settings of existing methods~\cite{chan2019everybody,wang2018video}; 2) \textbf{PL-Stage1-DA}: we add pose augmentation to PL-Stage1; 3) \textbf{PL-Stage1-DA-F}: based on PL-Stage1-DA, we change input poses to $\{\mathbf{p}_{t-K},\ldots,\mathbf{p}_{t+K}\}$; 4) \textbf{PL-Stage2}: we add the second refinement stage to PL-Stage1-DA-F, but still without unpaired learning; 5) \textbf{PL-UL-Stage2}: our full method, with unpaired learning used on both stages.

Table~\ref{tab:ablation} summarizes the quantitative ablation analysis results. We can see that by incrementally introducing these key components, both SSIM and FID metrics gradually improve supporting that for photorealistic human motion retargeting all the proposed contributions are essential. We also present visual results for ablation analysis in Figure~\ref{fig:ablation}. By zooming in certain areas, we can clearly find that comparing with the baseline, our full method is able to produce better mask boundaries, fewer artifacts, and richer and sharper texture details.


\begin{figure}[t]
\centering
\begin{subfigure}{0.48\linewidth}
\centering
  \includegraphics[width=1\linewidth]{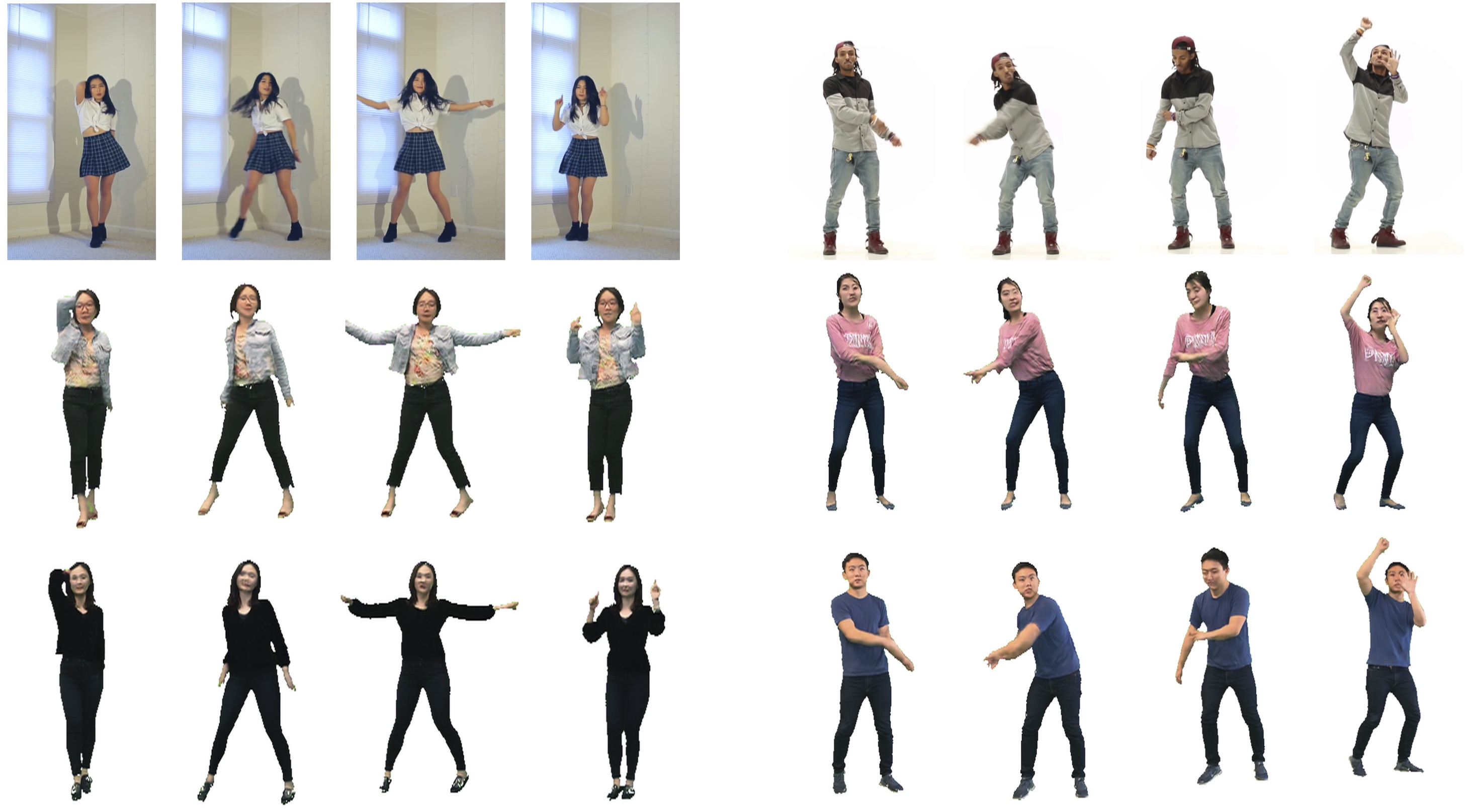}
  \caption{Single frame results.}\label{fig:demo1}
\end{subfigure} \hfill
\begin{subfigure}{0.48\linewidth}
\centering
  \includegraphics[width=1\linewidth]{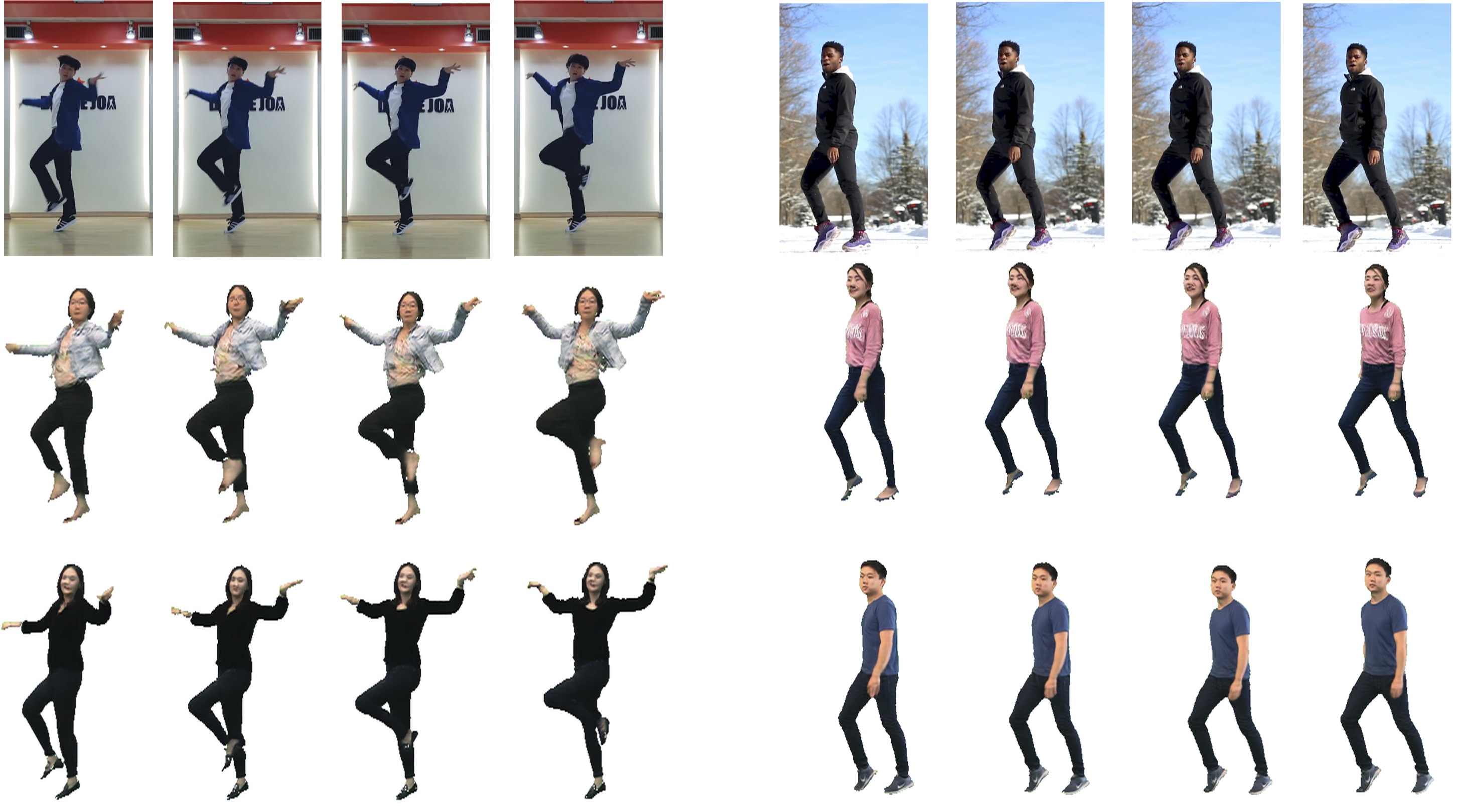}
  \caption{Consecutive frame results.}\label{fig:demo2}
\end{subfigure}
\caption{\textbf{Motion transfer results}. Both (a) single and (b) consecutive frames results are shown here. In each example, the first row shows the reference video, and the other rows show the transferred results on two different target persons.} \label{fig:demo}
\end{figure}

\subsection{Qualitative Results}
More motion transfer examples generated by our method are shown in Figure~\ref{fig:demo}, including results for both uniformly sampled single frames and consecutive frame sequences. These various results demonstrate that our method can generate high-fidelity results on various subjects from challenging body pose sequences.

\section{Conclusions}
 We introduced a novel approach for video human motion transfer. The network contains two stages with one sub-network for pose-to-image translation and another for image-to-image translation. The network is trained by stacked continuous frames to achieve temporally consistent results. We also incorporate a unified paired and unpaired learning strategy and a pose augmentation method during training to help the network generalize well on unseen in-the-wild poses. Our experiments support that all the proposed contributions are essential for obtaining realistic human motion retargteting. To further boost research on the topic, we collected two datasets containing human motion videos. The first dataset is used for training personalized models, while the second one for unpaired learning and evaluation.

\clearpage
%
%
\bibliographystyle{splncs04}
\bibliography{egbib}

\begin{thebibliography}{10}
\providecommand{\url}[1]{\texttt{#1}}
\providecommand{\urlprefix}{URL }
\providecommand{\doi}[1]{https://doi.org/#1}

\bibitem{aberman2019deep}
Aberman, K., Shi, M., Liao, J., Liscbinski, D., Chen, B., Cohen-Or, D.: Deep
  video-based performance cloning. In: Computer Graphics Forum. vol.~38, pp.
  219--233. Wiley Online Library (2019)

\bibitem{alp2018densepose}
Alp~G{\"u}ler, R., Neverova, N., Kokkinos, I.: Densepose: Dense human pose
  estimation in the wild. In: Proceedings of the IEEE Conference on Computer
  Vision and Pattern Recognition. pp. 7297--7306 (2018)

\bibitem{elor2017bringingPortraits}
Averbuch-Elor, H., Cohen-Or, D., Kopf, J., Cohen, M.F.: Bringing portraits to
  life. ACM Transactions on Graphics (Proceeding of SIGGRAPH Asia 2017)
  \textbf{36}(6), ~196 (2017)

\bibitem{balakrishnan2018synthesizing}
Balakrishnan, G., Zhao, A., Dalca, A.V., Durand, F., Guttag, J.: Synthesizing
  images of humans in unseen poses. In: Proceedings of the IEEE Conference on
  Computer Vision and Pattern Recognition. pp. 8340--8348 (2018)

\bibitem{bansal2018recycle}
Bansal, A., Ma, S., Ramanan, D., Sheikh, Y.: Recycle-gan: Unsupervised video
  retargeting. In: Proceedings of the European Conference on Computer Vision
  (ECCV). pp. 119--135 (2018)

\bibitem{de2018semi}
de~Bem, R., Ghosh, A., Ajanthan, T., Miksik, O., Siddharth, N., Torr, P.: A
  semi-supervised deep generative model for human body analysis. In:
  Proceedings of the European Conference on Computer Vision (ECCV). pp.~0--0
  (2018)

\bibitem{cao2018openpose}
Cao, Z., Hidalgo, G., Simon, T., Wei, S.E., Sheikh, Y.: Openpose: realtime
  multi-person 2d pose estimation using part affinity fields. arXiv preprint
  arXiv:1812.08008  (2018)

\bibitem{chan2019everybody}
Chan, C., Ginosar, S., Zhou, T., Efros, A.A.: Everybody dance now. In:
  Proceedings of the IEEE International Conference on Computer Vision. pp.
  5933--5942 (2019)

\bibitem{chen2017coherent}
Chen, D., Liao, J., Yuan, L., Yu, N., Hua, G.: Coherent online video style
  transfer. In: Proceedings of the IEEE International Conference on Computer
  Vision. pp. 1105--1114 (2017)

\bibitem{deeplabv3plus2018}
Chen, L.C., Zhu, Y., Papandreou, G., Schroff, F., Adam, H.: Encoder-decoder
  with atrous separable convolution for semantic image segmentation. In: ECCV
  (2018)

\bibitem{denton2015deep}
Denton, E.L., Chintala, S., Fergus, R., et~al.: Deep generative image models
  using alaplacian pyramid of adversarial networks. In: Advances in neural
  information processing systems. pp. 1486--1494 (2015)

\bibitem{denton2017unsupervised}
Denton, E.L., et~al.: Unsupervised learning of disentangled representations
  from video. In: Advances in neural information processing systems. pp.
  4414--4423 (2017)

\bibitem{esser2019unsupervised}
Esser, P., Haux, J., Ommer, B.: Unsupervised robust disentangling of latent
  characteristics for image synthesis. In: Proceedings of the IEEE
  International Conference on Computer Vision. pp. 2699--2709 (2019)

\bibitem{esser2018variational}
Esser, P., Sutter, E., Ommer, B.: A variational u-net for conditional
  appearance and shape generation. In: Proceedings of the IEEE Conference on
  Computer Vision and Pattern Recognition. pp. 8857--8866 (2018)

\bibitem{fang2017rmpe}
Fang, H.S., Xie, S., Tai, Y.W., Lu, C.: Rmpe: Regional multi-person pose
  estimation. In: Proceedings of the IEEE International Conference on Computer
  Vision. pp. 2334--2343 (2017)

\bibitem{finn2016unsupervised}
Finn, C., Goodfellow, I., Levine, S.: Unsupervised learning for physical
  interaction through video prediction. In: Advances in neural information
  processing systems. pp. 64--72 (2016)

\bibitem{gafni2019vid2game}
Gafni, O., Wolf, L., Taigman, Y.: Vid2game: Controllable characters extracted
  from real-world videos. arXiv preprint arXiv:1904.08379  (2019)

\bibitem{ha2019marionette}
Ha, S., Kersner, M., Kim, B., Seo, S., Kim, D.: Marionette: Few-shot face
  reenactment preserving identity of unseen targets. arXiv preprint
  arXiv:1911.08139  (2019)

\bibitem{heusel2017gans}
Heusel, M., Ramsauer, H., Unterthiner, T., Nessler, B., Hochreiter, S.: Gans
  trained by a two time-scale update rule converge to a local nash equilibrium.
  In: Advances in Neural Information Processing Systems. pp. 6626--6637 (2017)

\bibitem{huang2017real}
Huang, H., Wang, H., Luo, W., Ma, L., Jiang, W., Zhu, X., Li, Z., Liu, W.:
  Real-time neural style transfer for videos. In: Proceedings of the IEEE
  Conference on Computer Vision and Pattern Recognition. pp. 783--791 (2017)

\bibitem{isola2017image}
Isola, P., Zhu, J.Y., Zhou, T., Efros, A.A.: Image-to-image translation with
  conditional adversarial networks. In: Proceedings of the IEEE conference on
  computer vision and pattern recognition. pp. 1125--1134 (2017)

\bibitem{johnson2016perceptual}
Johnson, J., Alahi, A., Fei-Fei, L.: Perceptual losses for real-time style
  transfer and super-resolution. In: European conference on computer vision.
  pp. 694--711. Springer (2016)

\bibitem{joo2018generating}
Joo, D., Kim, D., Kim, J.: Generating a fusion image: One's identity and
  another's shape. In: Proceedings of the IEEE Conference on Computer Vision
  and Pattern Recognition. pp. 1635--1643 (2018)

\bibitem{kim2018deep}
Kim, H., Carrido, P., Tewari, A., Xu, W., Thies, J., Niessner, M., P{\'e}rez,
  P., Richardt, C., Zollh{\"o}fer, M., Theobalt, C.: Deep video portraits. ACM
  Transactions on Graphics (TOG)  \textbf{37}(4), ~163 (2018)

\bibitem{kim2019unsupervised}
Kim, Y., Nam, S., Cho, I., Kim, S.J.: Unsupervised keypoint learning for
  guiding class-conditional video prediction. In: Advances in Neural
  Information Processing Systems. pp. 3809--3819 (2019)

\bibitem{lee2019metapix}
Lee, J., Ramanan, D., Girdhar, R.: Metapix: Few-shot video retargeting. arXiv
  preprint arXiv:1910.04742  (2019)

\bibitem{liu2017unsupervised}
Liu, M.Y., Breuel, T., Kautz, J.: Unsupervised image-to-image translation
  networks. In: Advances in neural information processing systems. pp. 700--708
  (2017)

\bibitem{liu2019few}
Liu, M.Y., Huang, X., Mallya, A., Karras, T., Aila, T., Lehtinen, J., Kautz,
  J.: Few-shot unsupervised image-to-image translation. In: Proceedings of the
  IEEE International Conference on Computer Vision. pp. 10551--10560 (2019)

\bibitem{liu2019liquid}
Liu, W., Piao, Z., Min, J., Luo, W., Ma, L., Gao, S.: Liquid warping gan: A
  unified framework for human motion imitation, appearance transfer and novel
  view synthesis. In: Proceedings of the IEEE International Conference on
  Computer Vision. pp. 5904--5913 (2019)

\bibitem{lorenz2019unsupervised}
Lorenz, D., Bereska, L., Milbich, T., Ommer, B.: Unsupervised part-based
  disentangling of object shape and appearance. In: Proceedings of the IEEE
  Conference on Computer Vision and Pattern Recognition. pp. 10955--10964
  (2019)

\bibitem{ma2017pose}
Ma, L., Jia, X., Sun, Q., Schiele, B., Tuytelaars, T., Van~Gool, L.: Pose
  guided person image generation. In: Advances in Neural Information Processing
  Systems. pp. 406--416 (2017)

\bibitem{ma2018disentangled}
Ma, L., Sun, Q., Georgoulis, S., Van~Gool, L., Schiele, B., Fritz, M.:
  Disentangled person image generation. In: Proceedings of the IEEE Conference
  on Computer Vision and Pattern Recognition. pp. 99--108 (2018)

\bibitem{martin2018lookingood}
Martin-Brualla, R., Pandey, R., Yang, S., Pidlypenskyi, P., Taylor, J.,
  Valentin, J., Khamis, S., Davidson, P., Tkach, A., Lincoln, P., et~al.:
  Lookingood: enhancing performance capture with real-time neural re-rendering.
  arXiv preprint arXiv:1811.05029  (2018)

\bibitem{mirza2014conditional}
Mirza, M., Osindero, S.: Conditional generative adversarial nets. arXiv
  preprint arXiv:1411.1784  (2014)

\bibitem{odena2017conditional}
Odena, A., Olah, C., Shlens, J.: Conditional image synthesis with auxiliary
  classifier gans. In: Proceedings of the 34th International Conference on
  Machine Learning-Volume 70. pp. 2642--2651. JMLR. org (2017)

\bibitem{radford2015unsupervised}
Radford, A., Metz, L., Chintala, S.: Unsupervised representation learning with
  deep convolutional generative adversarial networks. arXiv preprint
  arXiv:1511.06434  (2015)

\bibitem{saito2017temporal}
Saito, M., Matsumoto, E., Saito, S.: Temporal generative adversarial nets with
  singular value clipping. In: Proceedings of the IEEE International Conference
  on Computer Vision. pp. 2830--2839 (2017)

\bibitem{Siarohin_2019_NeurIPS}
Siarohin, A., Lathuilière, S., Tulyakov, S., Ricci, E., Sebe, N.: First order
  motion model for image animation. In: Conference on Neural Information
  Processing Systems (NeurIPS) (December 2019)

\bibitem{siarohin2018deformable}
Siarohin, A., Sangineto, E., Lathuili{\`e}re, S., Sebe, N.: Deformable gans for
  pose-based human image generation. In: Proceedings of the IEEE Conference on
  Computer Vision and Pattern Recognition. pp. 3408--3416 (2018)

\bibitem{simonyan2014very}
Simonyan, K., Zisserman, A.: Very deep convolutional networks for large-scale
  image recognition. arXiv preprint arXiv:1409.1556  (2014)

\bibitem{thies2016face2face}
Thies, J., Zollhofer, M., Stamminger, M., Theobalt, C., Nie{\ss}ner, M.:
  Face2face: Real-time face capture and reenactment of rgb videos. In:
  Proceedings of the IEEE Conference on Computer Vision and Pattern
  Recognition. pp. 2387--2395 (2016)

\bibitem{tulyakov2018mocogan}
Tulyakov, S., Liu, M.Y., Yang, X., Kautz, J.: Mocogan: Decomposing motion and
  content for video generation. In: Proceedings of the IEEE conference on
  computer vision and pattern recognition. pp. 1526--1535 (2018)

\bibitem{villegas2018neural}
Villegas, R., Yang, J., Ceylan, D., Lee, H.: Neural kinematic networks for
  unsupervised motion retargetting. In: Proceedings of the IEEE Conference on
  Computer Vision and Pattern Recognition. pp. 8639--8648 (2018)

\bibitem{villegas2017learning}
Villegas, R., Yang, J., Zou, Y., Sohn, S., Lin, X., Lee, H.: Learning to
  generate long-term future via hierarchical prediction. In: Proceedings of the
  34th International Conference on Machine Learning-Volume 70. pp. 3560--3569.
  JMLR. org (2017)

\bibitem{vondrick2016generating}
Vondrick, C., Pirsiavash, H., Torralba, A.: Generating videos with scene
  dynamics. In: Advances In Neural Information Processing Systems. pp. 613--621
  (2016)

\bibitem{walker2016uncertain}
Walker, J., Doersch, C., Gupta, A., Hebert, M.: An uncertain future:
  Forecasting from static images using variational autoencoders. In: European
  Conference on Computer Vision. pp. 835--851. Springer (2016)

\bibitem{walker2017pose}
Walker, J., Marino, K., Gupta, A., Hebert, M.: The pose knows: Video
  forecasting by generating pose futures. In: Proceedings of the IEEE
  International Conference on Computer Vision. pp. 3332--3341 (2017)

\bibitem{wang2018fewshotvid2vid}
Wang, T.C., Liu, M.Y., Tao, A., Liu, G., Kautz, J., Catanzaro, B.: Few-shot
  video-to-video synthesis. In: Advances in Neural Information Processing
  Systems (NeurIPS) (2019)

\bibitem{wang2018video}
Wang, T.C., Liu, M.Y., Zhu, J.Y., Liu, G., Tao, A., Kautz, J., Catanzaro, B.:
  Video-to-video synthesis. In: Advances in Neural Information Processing
  Systems (NeurIPS) (2018)

\bibitem{wang2018pix2pixHD}
Wang, T.C., Liu, M.Y., Zhu, J.Y., Tao, A., Kautz, J., Catanzaro, B.:
  High-resolution image synthesis and semantic manipulation with conditional
  gans. In: Proceedings of the IEEE Conference on Computer Vision and Pattern
  Recognition (2018)

\bibitem{wang2004image}
Wang, Z., Bovik, A.C., Sheikh, H.R., Simoncelli, E.P., et~al.: Image quality
  assessment: from error visibility to structural similarity. IEEE transactions
  on image processing  \textbf{13}(4),  600--612 (2004)

\bibitem{wiles2018x2face}
Wiles, O., Sophia~Koepke, A., Zisserman, A.: X2face: A network for controlling
  face generation using images, audio, and pose codes. In: Proceedings of the
  European Conference on Computer Vision (ECCV). pp. 670--686 (2018)

\bibitem{xiu2018pose}
Xiu, Y., Li, J., Wang, H., Fang, Y., Lu, C.: Pose flow: Efficient online pose
  tracking. arXiv preprint arXiv:1802.00977  (2018)

\bibitem{yi2017dualgan}
Yi, Z., Zhang, H., Tan, P., Gong, M.: Dualgan: Unsupervised dual learning for
  image-to-image translation. In: Proceedings of the IEEE international
  conference on computer vision. pp. 2849--2857 (2017)

\bibitem{zakharov2019few}
Zakharov, E., Shysheya, A., Burkov, E., Lempitsky, V.: Few-shot adversarial
  learning of realistic neural talking head models. In: Proceedings of the IEEE
  International Conference on Computer Vision. pp. 9459--9468 (2019)

\bibitem{zanfir2018human}
Zanfir, M., Popa, A.I., Zanfir, A., Sminchisescu, C.: Human appearance
  transfer. In: Proceedings of the IEEE Conference on Computer Vision and
  Pattern Recognition. pp. 5391--5399 (2018)

\bibitem{zhang2018unreasonable}
Zhang, R., Isola, P., Efros, A.A., Shechtman, E., Wang, O.: The unreasonable
  effectiveness of deep features as a perceptual metric. In: Proceedings of the
  IEEE Conference on Computer Vision and Pattern Recognition. pp. 586--595
  (2018)

\bibitem{zhou2019dance}
Zhou, Y., Wang, Z., Fang, C., Bui, T., Berg, T.L.: Dance dance generation:
  Motion transfer for internet videos. arXiv preprint arXiv:1904.00129  (2019)

\bibitem{zhu2017unpaired}
Zhu, J.Y., Park, T., Isola, P., Efros, A.A.: Unpaired image-to-image
  translation using cycle-consistent adversarial networks. In: Proceedings of
  the IEEE international conference on computer vision. pp. 2223--2232 (2017)

\end{thebibliography}
\end{document}